\documentclass[twoside,11pt]{article}

\usepackage{jmlr2earxiv}

\usepackage[utf8]{inputenc} % allow utf-8 input
\usepackage[T1]{fontenc}    % use 8-bit T1 fonts
\usepackage[english]{babel}
\usepackage{booktabs,arydshln}
\usepackage{pycode}

\ShortHeadings{The \texttt{autofeat} Python Library}{Horn, Pack, and Rieger}
\firstpageno{1}

\begin{document}

\title{The \texttt{autofeat} Python Library for Automated\\ Feature Engineering and Selection}

\author{\name Franziska Horn \email cod3licious@gmail.com \\
       \addr Technische Universität Berlin\\
       Machine Learning Group\\
       Marchstr.\ 23, 10587 Berlin, Germany
       \AND
       \name Robert Pack \email robert.pack@basf.com\\
       \name Michael Rieger \email michael.rieger@basf.com \\
       \addr BASF SE\\
       Carl-Bosch-Str.\ 38, 67056 Ludwigshafen, Germany}

\maketitle

\begin{abstract}%   <- trailing '%' for backward compatibility of .sty file
This paper describes the \texttt{autofeat} Python library, which provides \texttt{scikit-learn} style linear regression and classification models with automated feature engineering and selection capabilities. Complex non-linear machine learning models, such as neural networks, are in practice often difficult to train and even harder to explain to non-statisticians, who require transparent analysis results as a basis for important business decisions. While linear models are efficient and intuitive, they generally provide lower prediction accuracies. Our library provides a multi-step feature engineering and selection process, where first a large pool of non-linear features is generated, from which then a small and robust set of meaningful features is selected, which improve the prediction accuracy of a linear model while retaining its interpretability.
\end{abstract}


\section{Introduction}
More and more companies aim to improve production processes with data science and machine learning (ML) methods, for example, by using a ML model to better understand which factors contribute to higher quality products or greater production yield. While advanced ML models such as neural networks (NN) might, theoretically, in many cases provide the most accurate predictions, they have several drawbacks in practice. First of all, with many hyperparameters to set, these model can be difficult and time consuming to fit, which is only aggravated by the current shortage of ML specialists in industry. Second, in many cases there is not enough data available in the first place to train a low bias/high variance model like a NN, for example, because comprehensive data collection pipelines are not yet fully implemented or because obtaining individual data points is expensive, e.g., when it takes several days to produce a single product. Last but not least, the insights generated by a ML analysis need to be communicated to others in the company, who want to use these results as a basis for important business decisions \cite{lime}. While great progress has been made to improve the interpretability of NNs, e.g., by using layer-wise relevance propagation (LRP) to reveal which of the input features contributed most to a neural net's prediction \cite{arras2017relevant,bach2015pixel,montavon2018methods}, this is in practice still not sufficient to convince those with only a limited understanding of statistics. Especially when dealing with data collected from physical systems, using a plausible model might even be more important than getting small prediction errors \cite{martius2016extrapolation}.

To avoid these shortcomings of NNs and other non-linear ML models, in practice we find it necessary to rely mostly on linear prediction models, which are intuitive to understand and can be trained easily and efficiently even on very small datasets.
But of course, employing linear models generally comes at the cost of a lower prediction accuracy, because in most datasets there is no linear relation between the original input features and the target variable.

The ability to learn expressive representations from the given input features is one of the main reasons for the popularity of neural networks and ``deep learning'' \cite{bengio2013deep,lecun2015deep}. Upon closer examination, a NN is nothing more than a linear model operating on better features: While a complex layer hierarchy maps the inputs to the last hidden layer, thereby transforming the original features into a more informative representation, the output layer, acting on the activations of this last hidden layer to generate the final prediction, corresponds to a simple linear model. Using pre-trained NNs as feature extractors, i.e., to transform the original inputs into more useful representations, often improves the performance of simpler models on tasks such as image classification \cite{sharif2014cnn}. To improve forecasts involving time series data, echo state networks use a randomly connected ``reservoir'' to create more informative feature vectors that are then used to train a ridge regression model \cite{lukovsevivcius2012practical}.
Similarly, kernel methods like SVM use the kernel trick to employ linear models but implicitly operate in a very high dimensional feature space where, e.g., classification problems become linearly separable \cite{muller2001introduction}. As these examples demonstrate, linear models are very capable of solving complex problems -- provided the right features are available. While NNs and kernel methods transform the original inputs into more useful feature representations internally, explicit feature engineering aims to create better features in a preprocessing step, i.e., before using the data to fit a (linear) prediction model.

Manually engineering new, more informative features is often quite tedious. Therefore, inspired by the SISSO algorithm \cite{ouyang2018sisso}, we propose a framework to automatically generate several tens of thousands of non-linear features from the original inputs and then carefully select the most informative of them as additional input features for a linear model. We have found that this approach leads to sufficiently accurate predictions on real world data while providing a transparent model that has a high acceptance rate amongst non-statisticians in the company and therefore provides the possibility to positively contribute to important business decisions.
To make this framework more accessible to other data scientists, our implementation is publicly available on GitHub.\footnote{\url{https://github.com/cod3licious/autofeat}}

The rest of the paper is structured as follows: After introducing some related work in the area of automated feature engineering and selection, we describe our approach and the \texttt{autofeat} Python library in detail (Section~\ref{sec:autofeat}). We then report experimental results on several datasets (Section~\ref{sec:exp}) before concluding the paper with a brief discussion (Section~\ref{sec:discussion}).

\subsection{Related Work}
Feature construction frameworks generally include both a feature engineering, as well as a feature selection component \cite{markovitch2002feature}. One of the main differences between feature construction approaches is whether they first generate an exhaustive feature pool and then perform feature selection on the whole feature set (which is also the strategy \texttt{autofeat} follows), or if the set of features is expanded iteratively, by evaluating at each step whether the inclusion of the new features would improve the prediction accuracy. Both approaches have their drawbacks: The first approach is very memory intensive, especially when starting off with a large initial feature set from which the additional features are constructed via various transformations. With the second approach, important features might be missed if some variables are eliminated too early in the feature engineering process and can therefore not serve to construct more complex, possibly helpful features. Furthermore, depending on the strategy for including additional features, the whole process might either be very time intensive, if at each step a model is trained and evaluated on the feature subset, or can fail to include (only) the relevant features, if a simple heuristic is used for the feature evaluation and selection.

Most existing feature construction frameworks follow the second, iterative feature engineering approach: The FICUS algorithm \cite{markovitch2002feature} uses a beam search to expand the feature space based on a simple heuristic, while the FEADIS algorithm \cite{dor2012strengthening} and Cognito \cite{khurana2016cognito} use more complex selection strategies. A more recent trend is to use meta-learning, i.e., algorithms trained on other datasets, to decide whether to apply specific transformation to the features or not \cite{katz2016explorekit,khurana2018feature,nargesian2017learning}. While theoretically promising, we could not find an easy to use open source library for any of these approaches, which makes them essentially irrelevant for practical data science use cases.

The well-known \texttt{scikit-learn} Python library \cite{scikit-learn} provides a function to generate polynomial features (e.g.\ $x^2$), including feature interactions (e.g.\ $x_1\cdot x_2, x_1^2\cdot x_2^3$). Polynomial features are a subset of the features generated by \texttt{autofeat}, yet, while they might be helpful for many datasets, in our experience with \texttt{autofeat}, a lot of times the ratios of two features or feature combinations turn out to be informative additional features, which can not be generated with the \texttt{scikit-learn} method.
The \texttt{scikit-learn} library also contains several options for feature selection, such as univariate feature scoring, recursive feature elimination, and other model-based feature selection approaches \cite{guyon2003introduction,kursa2010feature}. Univariate feature selection methods consider each feature individually, which can lead to the inclusion of many correlated features, like those contained in the feature pool generated by \texttt{autofeat}. The more sophisticated feature selection techniques rely on the use of an external prediction model that provides coefficients indicating the importance of each feature. However, algorithms such as linear regression get numerically unstable if the number of features is larger than the number of samples, which makes these approaches impractical for feature pools as large as those generated by \texttt{autofeat}.

One popular Python library for automated feature engineering is \texttt{featuretools}, which generates a large feature set using ``deep feature synthesis'' \cite{kanter2015deep}. This library is targeted towards relational data, where features can be created through aggregations (e.g.\ given some customers (data table 1) and their associated loans (in table 2), a new feature could be the sum of each customer's loans), or transformations (e.g.\ time since the last loan payment). A similar approach is also implemented by the ``one button machine'' \cite{lam2017one}. The strategy followed by \texttt{autofeat} is somewhat orthogonal to that of \texttt{featuretools}: It is not meant for relational data, found in many business application areas, but was rather built with scientific use cases in mind, where e.g.\ experimental measurements would instead be stored in a single table. For this reason, \texttt{autofeat} also makes it possible to specify the units of the input variables to prevent the creation of physically nonsensical features.

Another Python library worth mentioning is \texttt{tsfresh} \cite{christ2018time,christ2016distributed}, which provides feature engineering methods for time series, together with a univariate feature selection strategy. However, while \texttt{autofeat} can be applied to a variety of datasets, the features generated by \texttt{tsfresh} only make sense for time series data, as they are constructed, e.g., using rolling windows.

To the best of our knowledge, there does not exist a general purpose open source library for automated feature engineering and selection, which is why we felt compelled to share our work.

\section{Automated Feature Engineering and Selection with \texttt{autofeat}}\label{sec:autofeat}
The \texttt{autofeat} library provides the \texttt{AutoFeatRegressor} and \texttt{AutoFeatClassifier} models, which automatically generate and select additional non-linear input features given the original data and then train a linear prediction model with these features. The models provide a familiar \texttt{scikit-learn} \cite{scikit-learn} style interface, as demonstrated by a simple usage example, where \texttt{X} corresponds to a $n \times d$ feature matrix and \texttt{y} to an $n$-dimensional target vector (both NumPy arrays \cite{numpy} and Pandas DataFrames \cite{pandas} are supported as inputs): \vspace{-0.5cm}
\begin{lstlisting}
# instantiate the model
model = AutoFeatRegressor()
# fit the model and get a pandas DataFrame with the original,
# as well as the additional non-linear features
df = model.fit_transform(X, y)
# predict the target for new test data points
y_pred = model.predict(X_test)
# compute the additional features for new test data points
# (e.g. as input for a different model)
df_test = model.transform(X_test)
\end{lstlisting}

In the following, we describe the feature engineering and selection steps happening during a call to e.g.\ \texttt{AutoFeatRegressor.fit()} or \texttt{AutoFeatRegressor.fit\_transform()} in more detail. The \texttt{autofeat} library requires Python 3 and is pip-installable.

\subsection{Construction of Non-Linear Features}
Additional non-linear features are generated in an alternating multi-step process by applying user selectable non-linear transformations to the features (e.g.\ $\log(x)$, $\sqrt{x}$, $1/x$, $x^2$, $x^3$, $|x|$, $\exp(x)$, $2^x$, $\sin(x)$, $\cos(x)$) and combining pairs of features with different operators ($+, -, \cdot$). This results in an exponentially growing feature space, e.g., with only three original features, the first feature engineering step (applying non-linear transformation) results in about 20 new features, the second step (combining features), results in about 750 new features, and after a third step (again applying transformations), the feature space has grown to include over 4000 features. As this may require a fair amount of RAM depending on the number of original input features, the data points can be subsampled before computing the new features. In practice, performing only two or three feature engineering steps is usually sufficient.

The new features are computed using the SymPy Python library \cite{sympy}, which automatically simplifies the generated mathematical expressions and thereby makes it possible to exclude redundant features. If the original features are provided with physical units, only `legal' new features are retained, e.g., a feature representing a temperature would not be subtracted from a feature representing a volume of something. This is implemented using the Pint Python library,\footnote{\url{https://pint.readthedocs.io/en/latest/}} which is additionally used to compute several dimensionless quantities from the original features using the Buckingham $\pi$-theorem \cite{buckingham1914physically}. If categorical features are included in the original features, these are first transformed into one-hot encoded vectors using the corresponding \texttt{scikit-learn} model before using them in the main feature engineering procedure.

\subsection{Feature Selection}
After having generated several thousands of features (often more than data points in the original dataset), it is now indispensable to carefully select only those features that contribute meaningful information when used as input to a linear model. To this end, we first remove those engineered features that are highly correlated with the original or other simpler features and then employ a multi-step feature selection approach relying heavily on L1-regularized linear models. In addition to the \texttt{AutoFeatRegressor} and \texttt{AutoFeatClassifier} models, the library also provides only this feature selection part alone in the \texttt{FeatureSelector} class, which again provides a \texttt{scikit-learn} style interface.

Individual features can provide redundant information or they might seem uninformative by themselves yet proof useful in combination with others. Therefore, instead of ranking the features independently by some criterion, it is advantageous to use a wrapper method that considers multiple features at once to select a promising subset \cite{guyon2003introduction}. For this we use the Lasso LARS regression model \cite{baraniuk2007compressive,efron2004least,friedman2010regularization} and an L1-regularized logistic regression model \cite{cox1958regression,bishop} provided in the \texttt{scikit-learn} library, which yield sparse weights based on which the features can be chosen \cite{ng2004feature}. To select the features, we mainly rely on a noise filtering approach, where the model is trained on the original features, as well as several additional `noise' features (either created by shuffling the original data or randomly drawn from a normal distribution), and only those of the original features are kept that have a model coefficient larger than the largest coefficient associated with any of the noise features \cite{kursa2010feature}.

Selecting relevant features with an L1-regularized model amongst a feature pool that contains more features than data samples works quite well when the features are independent \cite{ng2004feature,doquetagnostic}. However, when trained with a large set of interrelated features, such as those generated in our feature engineering process, the models often fail to identify all of the truly relevant features. Therefore, we first identify an initial set of promising features by training an L1-regularized linear model on all features and selecting those with the largest absolute coefficients. Then, the remaining features are split into equal chunks and combined with the initial set (such that each of the chunks contains less than $n/2$ features) and a model is then fit on each chunk to select additional features. The feature subsets are then combined and used to train another model based on which a final feature set is determined. To get a more robust set of features, this selection process is performed multiple times on subsamples of the data. The feature subsets of the independent feature selection runs are then combined and highly correlated features are filtered out (keeping those features that were selected in the most runs). The remaining features are then again used to fit a model to select the ultimate feature set.

After this multi-step selection process, typically only a few dozen of the several thousand engineered features are retained and used to train the actual prediction model. For new test data points, the \texttt{AutoFeatRegressor} and \texttt{AutoFeatClassifier} models can then either generate predictions directly, or a DataFrame with the new features can be computed for all data points and used to train other models.
By examining the coefficients of the linear prediction model (possibly normalized by the standard deviation of the corresponding features, in case these are not of comparable magnitudes), the most prominent influencing factors related to higher or lower values of the target variable can be identified.

\section{Experimental Results}\label{sec:exp}
To give an indication of the performance of the \texttt{AutoFeatRegressor} model in practice, compared to other non-linear ML algorithms, we test our approach on five regression datasets (Table~\ref{table:datasets}), provided in the \texttt{scikit-learn} package (\emph{diabetes} and \emph{boston}) or obtainable from the UCI Machine Learning Repository.\footnote{\url{http://archive.ics.uci.edu/ml/index.php}} For further details on the experiments, including the hyperparameter selection of the other models, please refer to the corresponding Jupyter notebook in the GitHub repository.
\begin{table}[!htb]
\centering\setlength{\tabcolsep}{5pt}
\caption{Overview of datasets, including the number of samples $n$ and number of original input features $d$.}
\begin{tabular}{lr r l}
\toprule
Dataset             & $n$ & $d$ & Prediction task\\\midrule
\emph{diabetes} \cite{efron2004least}         & 442 & 10 & disease progression one year after baseline\\
\emph{boston} \cite{harrison1978hedonic}      & 506 & 13 & median housing values in suburbs of Boston\\
\emph{concrete} \cite{yeh1998modeling}        & 1030 & 8 & compressive strengths of concrete mixtures\\
\emph{airfoil} \cite{brooks1989airfoil}       & 1503 & 5 & sound pressure levels of airfoils in a wind tunnel\\
\emph{wine quality} \cite{cortez2009modeling} & 6497 & 12 & red \& white wine quality from physiochemical tests\\
\bottomrule
\end{tabular}
\label{table:datasets}
\end{table}

While on most datasets, the \texttt{AutoFeatRegressor} model does not quite reach the state-of-the-art performance of a random forest regression model (Table~\ref{table:results}), it clearly outperforms standard linear ridge regression, while retaining its interpretability. Across all datasets, with one feature engineering step, \texttt{autofeat} generated between 2 and 11 additional features, while with two and three steps, it produced on average 31 additional features (Table~\ref{table:nfeat}). Most of the selected features are ratios or products of (transformed) features (Table~\ref{table:feattypes}).

\begin{table}[!htb]
\centering\setlength{\tabcolsep}{3pt}
\caption{$R^2$ scores on the training and test folds of different datasets for ridge regression (RR), support vector regression (SVR), random forests (RF), and the \texttt{autofeat} regression model with one, two, or three feature engineering steps (AFR1-3). Best results per column are in boldface (existing methods) and underlined (AFR).}\small
\begin{tabular}{lcccccccccc}
\toprule
& \multicolumn{2}{c}{\textbf{diabetes}}& \multicolumn{2}{c}{\textbf{boston}}& \multicolumn{2}{c}{\textbf{concrete}}& \multicolumn{2}{c}{\textbf{airfoil}}& \multicolumn{2}{c}{\textbf{wine quality}}\\ \cmidrule(l){2-3}\cmidrule(l){4-5}\cmidrule(l){6-7}\cmidrule(l){8-9}\cmidrule(l){10-11}
            & train & test & train & test & train & test & train & test & train & test\\\midrule
\emph{RR}   & 0.541 & \textbf{0.383}& 0.736 & 0.748& 0.625 & 0.564& 0.517 & 0.508& 0.293 & 0.310\\
\emph{SVR}  & 0.580 & 0.320& 0.959 & \textbf{0.882}& 0.933 & 0.881& 0.884 & 0.851& 0.572 & 0.411\\
\emph{RF}   & \textbf{0.598} & 0.354& \textbf{0.983} & 0.870& \textbf{0.985} & \textbf{0.892}& \textbf{0.991} & \textbf{0.934}& \textbf{0.931} & \textbf{0.558}\\\addlinespace[0.5ex]
\hdashline \addlinespace[0.5ex]
\emph{AFR1} & 0.553 & \underline{0.400}& 0.825 & \underline{0.810}& 0.847 & 0.818& 0.569 & 0.560& 0.320 & 0.341\\
\emph{AFR2} & 0.591 & 0.353& 0.893 & 0.791& \underline{0.913} & \underline{0.868}& 0.863 & 0.842& \underline{0.397} & \underline{0.384}\\
\emph{AFR3} & \underline{0.638} & -12.4& \underline{0.932} & 0.048& 0.867 & 0.824& \underline{0.884} & \underline{0.883}& 0.350 & 0.342\\
\bottomrule
\end{tabular}
\label{table:results}
\end{table}

\begin{table}[!htb]
\centering\setlength{\tabcolsep}{5pt}
\caption{Number of engineered (eng) and selected (sel) additional features for each dataset from an \texttt{autofeat} regression model with one, two, or three feature engineering steps (AFR1-3).}
\begin{tabular}{lrrrrrrrrrr}
\toprule
& \multicolumn{2}{c}{\textbf{diabetes}}& \multicolumn{2}{c}{\textbf{boston}}& \multicolumn{2}{c}{\textbf{concrete}}& \multicolumn{2}{c}{\textbf{airfoil}}& \multicolumn{2}{c}{\textbf{wine quality}}\\ \cmidrule(l){2-3}\cmidrule(l){4-5}\cmidrule(l){6-7}\cmidrule(l){8-9}\cmidrule(l){10-11}
            & eng & sel & eng & sel & eng & sel & eng & sel & eng & sel\\\midrule
\emph{AFR1} & 45 & 2 & 60 & 6 & 34 & 5 & 21 & 6 & 59 & 11\\
\emph{AFR2} & 5950 & 8 & 10528 & 15 & 3456 & 40 & 530 & 42 & 9959 & 80\\
\emph{AFR3} & 32161 & 16 & 54631 & 21 & 14485 & 16 & 2355 & 44 & 55648 & 26\\
\bottomrule
\end{tabular}
\label{table:nfeat}
\end{table}
\begin{table}[!htb]
\centering\setlength{\tabcolsep}{5pt}
\caption{Most frequently selected features across all datasets for one, two, or three feature engineering steps (AFR1-3). Only the non-linear transformations $\log(x)$, $\sqrt{x}$, $1/x$, $x^2$, $x^3$, $|x|$, and $\exp(x)$ were applied during the feature engineering steps.}
\begin{tabular}{ll}
\toprule
\textbf{AFR1} & $1/x,\; x^3,\; x^2,\; \exp(x)$ \\[8pt]
\textbf{AFR2} & $\sqrt{x_1}/x_2,\; 1/(x_1x_2),\; x_1/x_2,\; x_1^3/x_2,\; x_1^2/x_2,\; \exp(x_1)\exp(x_2),\; \exp(x_1)/x_2,$\\[5pt]
              & $\sqrt{x_1}\sqrt{x_2},\; \sqrt{x_1}x_2^3,\; x_1\log(x_2),\; \log(x_1)/x_2,\; x_1^3x_2^3,\; x_1^3x_2,\; x_1^3\log(x_2), \; ...$\\[8pt]
\textbf{AFR3} & $x_1^3/x_2^3,\; \exp(\sqrt{x_1} - \sqrt{x_2}),\; 1/(x_1^3x_2^3),\; \sqrt{x_1x_2},\; 1/(x_1 + x_2),\; x_1/x_2^2,$\\[5pt]
              & $1/(\sqrt{x_1} - \log(x_2)),\; |\sqrt{x_1} - \log(x_2)|, \; \exp(\log(x_1)/x_2),\; \log(x_1)^2/x_2^2,$\\[5pt]
              & $ |\log(x_1) + \log(x_2)|,\; ...$\\
\bottomrule
\end{tabular}
\label{table:feattypes}
\end{table}

With only a single feature engineering step, the \texttt{AutoFeatRegressor} model often only performs slightly better than ridge regression on the original features. With three feature engineering steps, on the other hand, the model can overfit on the training data (as indicated by the discrepancy between the training and test $R^2$ scores), because the complex features do not only explain the signal, but also the noise contained in the data. However, the only datasets where this is a serious problem here is the \emph{diabetes} and \emph{boston} datasets, where over 30k and 50k features were generated in the feature engineering process, while less than 500 data points were available for feature selection and model fitting, which means overfitting is somewhat to be expected.

\section{Conclusion}\label{sec:discussion}
In this paper, we have introduced the \texttt{autofeat} Python library, which includes an automated feature engineering and selection procedure to improve the prediction accuracy of a linear model by using additional non-linear features. The regression and classification models are based on \texttt{scikit-learn} models and provide a familiar interface. During the model fit, a vast number of non-linear features is generated from the original features and a few of these are selected in an elaborate iterative process to optimally explain the target variable. By combining a linear model with complex non-linear features, a high prediction accuracy can be achieved, while retaining a transparent model that yields traceable results as a basis for business decisions made by non-statisticians.

The \texttt{autofeat} library was developed with scientific use cases in mind and is especially useful for heterogeneous datasets, e.g., containing sensor measurements with different physical units. It should not be seen as a competitor for the existing feature engineering libraries \texttt{featuretools} or \texttt{tsfresh}, which would be the first choice when dealing with relational business data or time series respectively.

We have demonstrated on several datasets that the \texttt{AutoFeatRegressor} model significantly improves upon the performance of a linear regression model and sometimes even outperforms other non-linear ML models. While the model can be used for predictions directly, it might also be beneficial to use the generated features as input to train other ML models. By adapting the kinds of transformations applied in the feature engineering process, as well as the number of feature engineering steps, further insights can be gained with respect to how which of the input features influences the target variable, as well as the complexity of the system as a whole.

% Acknowledgements should go at the end, before appendices and references
\acks{FH was a part-time employee at BASF when initially programming the \texttt{autofeat} library.}

\vskip 0.2in
\bibliography{horn19a}

\begin{thebibliography}{39}
\providecommand{\natexlab}[1]{#1}
\providecommand{\url}[1]{\texttt{#1}}
\expandafter\ifx\csname urlstyle\endcsname\relax
  \providecommand{\doi}[1]{doi: #1}\else
  \providecommand{\doi}{doi: \begingroup \urlstyle{rm}\Url}\fi

\bibitem[Arras et~al.(2017)Arras, Horn, Montavon, M{\"u}ller, and
  Samek]{arras2017relevant}
Leila Arras, Franziska Horn, Gr{\'e}goire Montavon, Klaus-Robert M{\"u}ller,
  and Wojciech Samek.
\newblock {``What is relevant in a text document?'': An interpretable machine
  learning approach}.
\newblock \emph{{PLOS ONE}}, 12\penalty0 (8):\penalty0 e0181142, 2017.

\bibitem[Bach et~al.(2015)Bach, Binder, Montavon, Klauschen, M{\"u}ller, and
  Samek]{bach2015pixel}
Sebastian Bach, Alexander Binder, Gr{\'e}goire Montavon, Frederick Klauschen,
  Klaus-Robert M{\"u}ller, and Wojciech Samek.
\newblock On pixel-wise explanations for non-linear classifier decisions by
  layer-wise relevance propagation.
\newblock \emph{{PLOS ONE}}, 10\penalty0 (7):\penalty0 e0130140, 2015.

\bibitem[Baraniuk(2007)]{baraniuk2007compressive}
Richard~G Baraniuk.
\newblock Compressive sensing [lecture notes].
\newblock \emph{IEEE Signal Processing Magazine}, 24\penalty0 (4):\penalty0
  118--121, 2007.

\bibitem[Bengio(2013)]{bengio2013deep}
Yoshua Bengio.
\newblock Deep learning of representations: Looking forward.
\newblock In \emph{Statistical language and speech processing}, pages 1--37.
  Springer, 2013.

\bibitem[Bishop(2006)]{bishop}
Christopher~M. Bishop.
\newblock \emph{Pattern Recognition and Machine Learning (Information Science
  and Statistics)}.
\newblock Springer-Verlag New York, Inc., Secaucus, NJ, USA, 2006.

\bibitem[Brooks et~al.(1989)Brooks, Pope, and Marcolini]{brooks1989airfoil}
Thomas~F Brooks, D~Stuart Pope, and Michael~A Marcolini.
\newblock Airfoil self-noise and prediction.
\newblock \emph{Technical report, NASA RP-1218}, 1989.

\bibitem[Buckingham(1914)]{buckingham1914physically}
Edgar Buckingham.
\newblock On physically similar systems; illustrations of the use of
  dimensional equations.
\newblock \emph{Physical review}, 4\penalty0 (4):\penalty0 345, 1914.

\bibitem[Christ et~al.(2016)Christ, Kempa-Liehr, and
  Feindt]{christ2016distributed}
Maximilian Christ, Andreas~W Kempa-Liehr, and Michael Feindt.
\newblock Distributed and parallel time series feature extraction for
  industrial big data applications.
\newblock \emph{arXiv preprint arXiv:1610.07717}, 2016.

\bibitem[Christ et~al.(2018)Christ, Braun, Neuffer, and
  Kempa-Liehr]{christ2018time}
Maximilian Christ, Nils Braun, Julius Neuffer, and Andreas~W Kempa-Liehr.
\newblock Time series feature extraction on basis of scalable hypothesis tests
  (tsfresh--a python package).
\newblock \emph{Neurocomputing}, 307:\penalty0 72--77, 2018.

\bibitem[Cortez et~al.(2009)Cortez, Cerdeira, Almeida, Matos, and
  Reis]{cortez2009modeling}
Paulo Cortez, Ant{\'o}nio Cerdeira, Fernando Almeida, Telmo Matos, and Jos{\'e}
  Reis.
\newblock Modeling wine preferences by data mining from physicochemical
  properties.
\newblock \emph{Decision Support Systems}, 47\penalty0 (4):\penalty0 547--553,
  2009.

\bibitem[Cox(1958)]{cox1958regression}
David~R Cox.
\newblock The regression analysis of binary sequences.
\newblock \emph{Journal of the Royal Statistical Society: Series B
  (Methodological)}, 20\penalty0 (2):\penalty0 215--232, 1958.

\bibitem[Doquet and Sebag()]{doquetagnostic}
Guillaume Doquet and Michele Sebag.
\newblock Agnostic feature selection.

\bibitem[Dor and Reich(2012)]{dor2012strengthening}
Ofer Dor and Yoram Reich.
\newblock Strengthening learning algorithms by feature discovery.
\newblock \emph{Information Sciences}, 189:\penalty0 176--190, 2012.

\bibitem[Efron et~al.(2004)Efron, Hastie, Johnstone, Tibshirani,
  et~al.]{efron2004least}
Bradley Efron, Trevor Hastie, Iain Johnstone, Robert Tibshirani, et~al.
\newblock Least angle regression.
\newblock \emph{The Annals of Statistics}, 32\penalty0 (2):\penalty0 407--499,
  2004.

\bibitem[Friedman et~al.(2010)Friedman, Hastie, and
  Tibshirani]{friedman2010regularization}
Jerome Friedman, Trevor Hastie, and Rob Tibshirani.
\newblock Regularization paths for generalized linear models via coordinate
  descent.
\newblock \emph{Journal of Statistical Software}, 33\penalty0 (1):\penalty0 1,
  2010.

\bibitem[Guyon and Elisseeff(2003)]{guyon2003introduction}
Isabelle Guyon and Andr{\'e} Elisseeff.
\newblock An introduction to variable and feature selection.
\newblock \emph{Journal of Machine Learning Research}, 3:\penalty0 1157--1182,
  2003.

\bibitem[Harrison~Jr and Rubinfeld(1978)]{harrison1978hedonic}
David Harrison~Jr and Daniel~L Rubinfeld.
\newblock Hedonic housing prices and the demand for clean air.
\newblock \emph{Journal of environmental economics and management}, 5\penalty0
  (1):\penalty0 81--102, 1978.

\bibitem[Kanter and Veeramachaneni(2015)]{kanter2015deep}
James~Max Kanter and Kalyan Veeramachaneni.
\newblock Deep feature synthesis: Towards automating data science endeavors.
\newblock In \emph{2015 {IEEE} International Conference on Data Science and
  Advanced Analytics, DSAA 2015, Paris, France, October 19-21, 2015}, pages
  1--10. IEEE, 2015.

\bibitem[Katz et~al.(2016)Katz, Shin, and Song]{katz2016explorekit}
Gilad Katz, Eui Chul~Richard Shin, and Dawn Song.
\newblock Explorekit: Automatic feature generation and selection.
\newblock In \emph{2016 IEEE 16th International Conference on Data Mining
  (ICDM)}, pages 979--984. IEEE, 2016.

\bibitem[Khurana et~al.(2016)Khurana, Turaga, Samulowitz, and
  Parthasrathy]{khurana2016cognito}
Udayan Khurana, Deepak Turaga, Horst Samulowitz, and Srinivasan Parthasrathy.
\newblock Cognito: Automated feature engineering for supervised learning.
\newblock In \emph{2016 IEEE 16th International Conference on Data Mining
  Workshops (ICDMW)}, pages 1304--1307. IEEE, 2016.

\bibitem[Khurana et~al.(2018)Khurana, Samulowitz, and
  Turaga]{khurana2018feature}
Udayan Khurana, Horst Samulowitz, and Deepak Turaga.
\newblock Feature engineering for predictive modeling using reinforcement
  learning.
\newblock In \emph{Thirty-Second {AAAI} Conference on Artificial Intelligence},
  2018.

\bibitem[Kursa et~al.(2010)Kursa, Rudnicki, et~al.]{kursa2010feature}
Miron~B Kursa, Witold~R Rudnicki, et~al.
\newblock Feature selection with the boruta package.
\newblock \emph{J Stat Softw}, 36\penalty0 (11):\penalty0 1--13, 2010.

\bibitem[Lam et~al.(2017)Lam, Thiebaut, Sinn, Chen, Mai, and Alkan]{lam2017one}
Hoang~Thanh Lam, Johann-Michael Thiebaut, Mathieu Sinn, Bei Chen, Tiep Mai, and
  Oznur Alkan.
\newblock One button machine for automating feature engineering in relational
  databases.
\newblock \emph{arXiv preprint arXiv:1706.00327}, 2017.

\bibitem[LeCun et~al.(2015)LeCun, Bengio, and Hinton]{lecun2015deep}
Yann LeCun, Yoshua Bengio, and Geoffrey Hinton.
\newblock Deep learning.
\newblock \emph{Nature}, 521\penalty0 (7553):\penalty0 436--444, 2015.

\bibitem[Luko{\v{s}}evi{\v{c}}ius(2012)]{lukovsevivcius2012practical}
Mantas Luko{\v{s}}evi{\v{c}}ius.
\newblock A practical guide to applying echo state networks.
\newblock In \emph{Neural networks: Tricks of the trade}, pages 659--686.
  Springer, 2012.

\bibitem[Markovitch and Rosenstein(2002)]{markovitch2002feature}
Shaul Markovitch and Dan Rosenstein.
\newblock Feature generation using general constructor functions.
\newblock \emph{Machine Learning}, 49\penalty0 (1):\penalty0 59--98, 2002.

\bibitem[Martius and Lampert(2016)]{martius2016extrapolation}
Georg Martius and Christoph~H Lampert.
\newblock Extrapolation and learning equations.
\newblock \emph{arXiv preprint arXiv:1610.02995}, 2016.

\bibitem[McKinney et~al.(2010)]{pandas}
Wes McKinney et~al.
\newblock Data structures for statistical computing in python.
\newblock In \emph{Proceedings of the 9th Python in Science Conference}, volume
  445, pages 51--56. Austin, TX, 2010.

\bibitem[Meurer et~al.(2017)Meurer, Smith, Paprocki, {\v{C}}ert{\'\i}k,
  Kirpichev, Rocklin, Kumar, Ivanov, Moore, Singh, et~al.]{sympy}
Aaron Meurer, Christopher~P Smith, Mateusz Paprocki, Ond{\v{r}}ej
  {\v{C}}ert{\'\i}k, Sergey~B Kirpichev, Matthew Rocklin, AMiT Kumar, Sergiu
  Ivanov, Jason~K Moore, Sartaj Singh, et~al.
\newblock Sympy: symbolic computing in python.
\newblock \emph{PeerJ Computer Science}, 3:\penalty0 e103, 2017.

\bibitem[Montavon et~al.(2018)Montavon, Samek, and
  M{\"u}ller]{montavon2018methods}
Gr{\'e}goire Montavon, Wojciech Samek, and Klaus-Robert M{\"u}ller.
\newblock Methods for interpreting and understanding deep neural networks.
\newblock \emph{Digital Signal Processing}, 73:\penalty0 1--15, 2018.

\bibitem[M{\"u}ller et~al.(2001)M{\"u}ller, Mika, R{\"a}tsch, Tsuda, and
  Sch{\"o}lkopf]{muller2001introduction}
Klaus-Robert M{\"u}ller, Sebastian Mika, Gunnar R{\"a}tsch, Koji Tsuda, and
  Bernhard Sch{\"o}lkopf.
\newblock An introduction to kernel-based learning algorithms.
\newblock \emph{IEEE Transactions on Neural Networks}, 12\penalty0
  (2):\penalty0 181--201, 2001.

\bibitem[Nargesian et~al.(2017)Nargesian, Samulowitz, Khurana, Khalil, and
  Turaga]{nargesian2017learning}
Fatemeh Nargesian, Horst Samulowitz, Udayan Khurana, Elias~B Khalil, and
  Deepak~S Turaga.
\newblock Learning feature engineering for classification.
\newblock In \emph{{IJCAI}}, pages 2529--2535, 2017.

\bibitem[Ng(2004)]{ng2004feature}
Andrew~Y Ng.
\newblock Feature selection, l1 vs. l2 regularization, and rotational
  invariance.
\newblock In \emph{Proceedings of the twenty-first international conference on
  Machine learning}, page~78. ACM, 2004.

\bibitem[Oliphant(2006)]{numpy}
Travis~E Oliphant.
\newblock \emph{A guide to NumPy}, volume~1.
\newblock Trelgol Publishing USA, 2006.

\bibitem[Ouyang et~al.(2018)Ouyang, Curtarolo, Ahmetcik, Scheffler, and
  Ghiringhelli]{ouyang2018sisso}
Runhai Ouyang, Stefano Curtarolo, Emre Ahmetcik, Matthias Scheffler, and Luca~M
  Ghiringhelli.
\newblock Sisso: A compressed-sensing method for identifying the best
  low-dimensional descriptor in an immensity of offered candidates.
\newblock \emph{Physical Review Materials}, 2\penalty0 (8):\penalty0 083802,
  2018.

\bibitem[Pedregosa et~al.(2011)Pedregosa, Varoquaux, Gramfort, Michel, Thirion,
  Grisel, Blondel, Prettenhofer, Weiss, Dubourg, et~al.]{scikit-learn}
Fabian Pedregosa, Ga{\"e}l Varoquaux, Alexandre Gramfort, Vincent Michel,
  Bertrand Thirion, Olivier Grisel, Mathieu Blondel, Peter Prettenhofer, Ron
  Weiss, Vincent Dubourg, et~al.
\newblock Scikit-learn: Machine learning in python.
\newblock \emph{Journal of Machine Learning Research}, 12\penalty0
  (Oct):\penalty0 2825--2830, 2011.

\bibitem[Ribeiro et~al.(2016)Ribeiro, Singh, and Guestrin]{lime}
Marco~Tulio Ribeiro, Sameer Singh, and Carlos Guestrin.
\newblock "why should {I} trust you?": Explaining the predictions of any
  classifier.
\newblock In \emph{Proceedings of the 22nd {ACM} {SIGKDD} International
  Conference on Knowledge Discovery and Data Mining, San Francisco, CA, USA,
  August 13-17, 2016}, pages 1135--1144, 2016.

\bibitem[Sharif~Razavian et~al.(2014)Sharif~Razavian, Azizpour, Sullivan, and
  Carlsson]{sharif2014cnn}
Ali Sharif~Razavian, Hossein Azizpour, Josephine Sullivan, and Stefan Carlsson.
\newblock Cnn features off-the-shelf: an astounding baseline for recognition.
\newblock In \emph{Proceedings of the IEEE conference on computer vision and
  pattern recognition workshops}, pages 806--813, 2014.

\bibitem[Yeh(1998)]{yeh1998modeling}
I-C Yeh.
\newblock Modeling of strength of high-performance concrete using artificial
  neural networks.
\newblock \emph{Cement and Concrete research}, 28\penalty0 (12):\penalty0
  1797--1808, 1998.

\end{thebibliography}

\end{document}